%% file: ijcai22.tex
\documentclass{article}
\usepackage{ijcai22}

\usepackage{times}
\usepackage{soul}
\usepackage{url}
\usepackage[hidelinks]{hyperref}
\usepackage[utf8]{inputenc}
\usepackage[small]{caption}
\usepackage{graphicx}
\usepackage{amsmath}
\usepackage{amsthm}
\usepackage{booktabs}
\usepackage{algorithm}
\usepackage{algorithmic}
\usepackage{makecell}

\input{math_commands.tex}

\usepackage{physics}
\usepackage{courier}
\newcommand{\ct}[1]{\texttt{#1}}

\usepackage{adjustbox}
\usepackage{multirow}
\usepackage{float}
\usepackage{array}
\newcolumntype{P}[1]{>{\centering\arraybackslash}p{#1}}
\newcolumntype{M}[1]{>{\centering\arraybackslash}m{#1}}

\title{Adapt to Adaptation: Learning Personalization for Cross-Silo Federated Learning}

\author{
Jun Luo$^1$
\and
Shandong Wu$^{1,2,3,4}$
\affiliations
$^1$Intelligent Systems Program, University of Pittsburgh\\
$^2$Department of Radiology, University of Pittsburgh\\
$^3$Department of Biomedical Informatics, University of Pittsburgh\\
$^4$Department of Bioengineering, University of Pittsburgh\\
\emails
jul117@pitt.edu,
wus3@upmc.edu
}

\begin{document}
\maketitle

\begin{abstract}
  Conventional federated learning (FL) trains one global model for a federation of clients with decentralized data, reducing the privacy risk of centralized training. However, the distribution shift across non-IID datasets, often poses a challenge to this one-model-fits-all solution. Personalized FL aims to mitigate this issue systematically. In this work, we propose \ct{APPLE}, a personalized cross-silo FL framework that adaptively learns how much each client can benefit from other clients’ models. We also introduce a method to flexibly control the focus of training \ct{APPLE} between global and local objectives. We empirically evaluate our method's convergence and generalization behaviors, and perform extensive experiments on two benchmark datasets and two medical imaging datasets under two non-IID settings. The results show that the proposed personalized FL framework, \ct{APPLE}, achieves state-of-the-art performance compared to several other personalized FL approaches in the literature. The code is publicly available at \url{https://github.com/ljaiverson/pFL-APPLE}.\footnote{An extended version of this paper (with the Appendix included) can be found at \url{https://arxiv.org/abs/2110.08394}.}
\end{abstract}

\section{Introduction}
In recent years, federated learning (FL)~\cite{mcmahan2017communication,kairouz2019advances} has shown great potential in training a shared global model for decentralized data. In contrast with previous large-scale machine learning approaches, training in FL resides on the sites of data owners, without the need to migrate the data, which reduces systemic privacy risks and expenses on massive datacenters~\cite{kairouz2019advances}. Compared to separate individual training, the leading FL algorithm, \ct{FedAvg}~\cite{mcmahan2017communication}, as a representative of global FL algorithms, attempts to train a consensus global model by iteratively averaging the local updates of the global model. However, such an approach often suffers from convergence challenges brought on by the statistical data heterogeneity~\cite{smith2017federated,li2020federated,hsieh2020non}, where data are not identically distributed (non-IID) across all clients due to the inherent diversity~\cite{sahu2018convergence,li2019convergence}.

Data heterogeneity lies almost everywhere in real-world FL applications. For instance, in a cross-device training of a mobile keyboard next-word prediction model, the data heterogeneity can be generated by different typing preferences of users~\cite{hard2018federated,yang2018applied}; in addition, medical datasets across different silos are heterogeneous by nature, due to factors such as different data acquisition protocols and various local demographics~\cite{rieke2020future}. Moreover, in cross-silo FL, data heterogeneity may lead to inferior performance of federated models in certain silos, which may cost their incentives of further participation in the federation.

\begin{figure}[t]
    \centering
    \includegraphics[width=8.2cm]{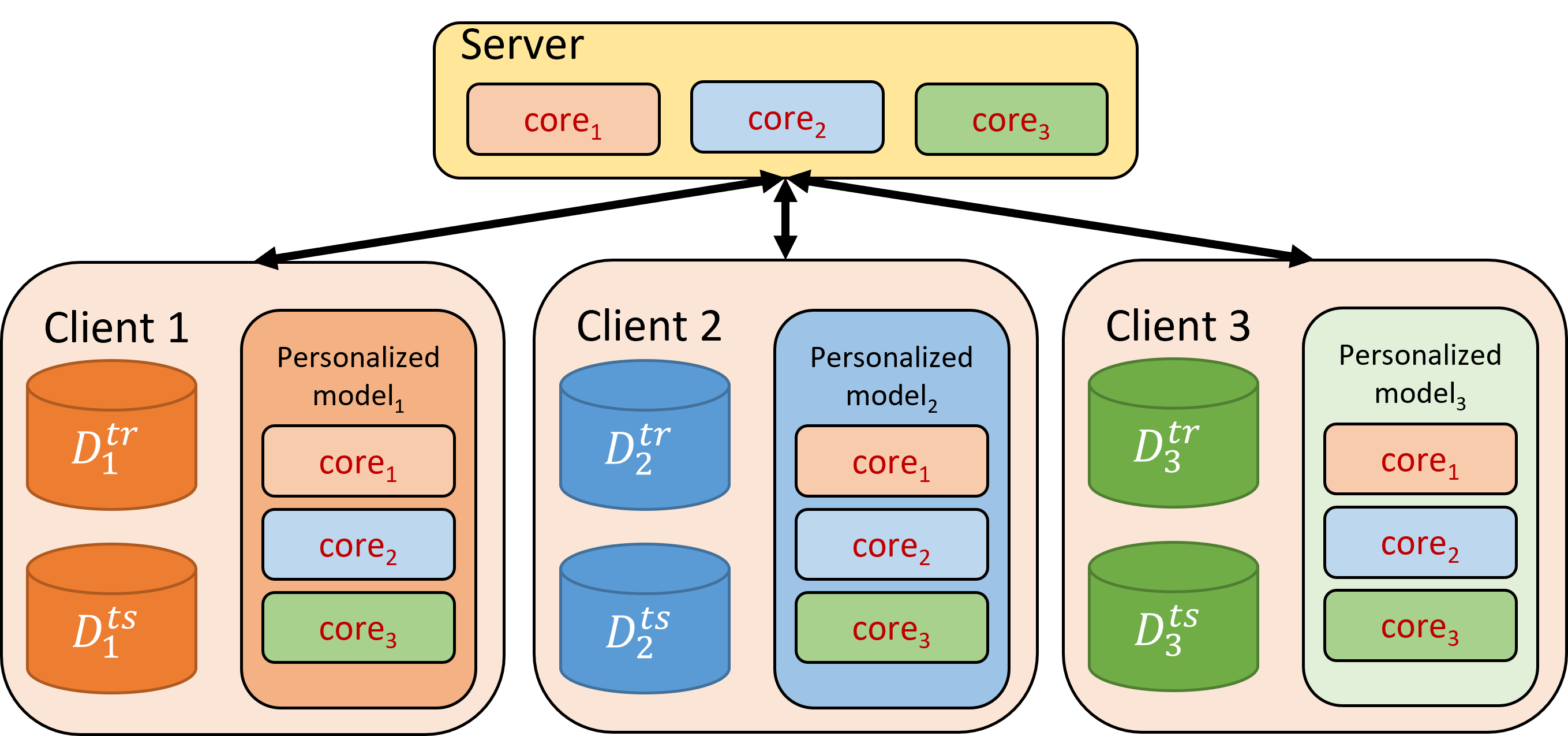}
    \caption{The workflow of the proposed framework, \ct{APPLE}. Each client trains a local personalized model, uploads its updated core model to the server, and downloads others' core models from the server as needed at the beginning of each round.} \label{apple-workflow}
\end{figure}

Based on the number of trained model(s), attempts to supplement FL algorithms with the ability to better handle data heterogeneity fall into two general schemes. The first scheme tries to enhance the global consensus model for higher robustness to non-IID datasets~\cite{li2020federated,karimireddy2020scaffold,acar2020federated,zhao2018federated}. The other scheme looks at FL from a client-centric perspective, aiming to train multiple models for different clients~\cite{kulkarni2020survey,kairouz2019advances}, and is often referred to as \textit{personalized} FL.

Personalized FL tries to systematically mitigate the influence of data heterogeneity, since a different model could be trained for a different target data distribution. Efforts in this direction include approaches that fine-tune the global model~\cite{wang2019federated}, and more sophisticated approaches that leverage meta-learning or multi-task learning~\cite{smith2017federated,sattler2020clustered} to learn the relationships between source and target domains/tasks, which corresponds to different distributions of the datasets. Other efforts pay more attention to interpolations between local models and the global model~\cite{deng2020adaptive,huang2021personalized,zhang2020personalized}.

In this work, we focus on the personalization aspect of cross-silo FL for non-IID data. We propose \textit{\textbf{A}da\textbf{p}tive \textbf{P}ersonalized Cross-Si\textbf{l}o F\textbf{e}derated Learning} (\ct{APPLE}), a novel personalized FL framework for cross-silo settings that adaptively learns to personalize each client's model by learning how much the client can benefit from other clients’ models, according to the local objective. In this process, the clients do not need to acquire information regarding  other clients' data distributions. We illustrate the workflow of \ct{APPLE} in Figure \ref{apple-workflow}.

There are three major distinctions between \ct{APPLE} and other existing personalized FL algorithms: in \ct{APPLE}, (1) after local training, a client does not upload the local personalized model, but a constructing component of the personalized model, here called a \textit{core model}; (2) the central server only maintains the core models uploaded from the clients, for further downloading purposes; (3) a unique set of local weights on each client, here called a \textit{directed relationship (DR)} vector, is adaptively learned to weight the downloaded core models from the central server. This enables the personalized models to take more advantage of the beneficial core models, while suppressing the less beneficial or potentially harmful core models. We also introduce a method to flexibly control the focus of training \ct{APPLE} between global and local objectives, by dynamically penalizing the DR vectors. The DR vectors are never uploaded to the server, nor shared with other clients. Coupled with uploading only the core models rather than the personalized models, the privacy risks are further reduced.

We summarize our contribution as follows:
\begin{itemize}
    \item We propose \ct{APPLE}, a novel personalized cross-silo FL framework that adaptively learns to personalize the client models.
    \item Within \ct{APPLE}, we introduce a method to flexibly control the focus of training between global and local objectives via a dynamic penalty.
    \item We evaluate \ct{APPLE} on two benchmark datasets and two medical imaging datasets under two types of non-IID settings. Our results show that \ct{APPLE} achieves state-of-the-art performance in both settings compared to other personalized FL approaches in the literature.
\end{itemize}

\section{Related Work}
\subsection{Federated Learning on Non-IID Data}
Federated learning (FL)~\cite{mcmahan2017communication,kairouz2019advances} enables participating clients to collaboratively train a model without migrating the clients' data, which mitigates the systemic privacy risks. The most notable FL algorithm, \ct{FedAvg}~\cite{mcmahan2017communication}, achieves this by aggregating the updated copies of the global model using an averaging approach. While concerns for the behavior of \ct{FedAvg} on non-IID data began to accumulate~\cite{sahu2018convergence}, numerous works have focused on the robustness of FL on non-IID data. Li et al.~\shortcite{li2020federated} proposed \ct{FedProx} that penalizes the local update when it is far from the prox-center. Karimireddy et al.~\shortcite{karimireddy2020scaffold} proposed \ct{SCAFFOLD} that corrects the local gradient under client-drift with control variates. \ct{FedDyn} by Acar et al.~\shortcite{acar2020federated} dynamically updates the regularizer in the empirical risk to reduce the impact of data heterogeneity.

\subsection{Personalized Federated Learning}
To systematically mitigate the impact of data heterogeneity, a new branch of FL, the \textit{personalized} FL, has emerged in recent years~\cite{kulkarni2020survey}. Instead of being restricted by the global consensus model, personalized FL allows different models for different clients, especially when the data are drawn from distinct distributions.

In this line of work, a natural way is to fine-tune the global model at each client~\cite{wang2019federated}. However, Jiang et al.~\shortcite{jiang2019improving} claimed that fine-tuning the global model may result in poor generalization to unseen data. They also demonstrated the similarity between personalizing a \ct{FedAvg}-trained model and meta-learning~\cite{finn2017model}. Some recent works have focused on the overlap between meta-learning and FL~\cite{fallah2020personalized,khodak2019adaptive}.


Apart from approaches that need further fine-tuning the trained models, Smith et al.~\shortcite{smith2017federated} proposed \ct{MOCHA} that leverages multi-task learning to learn the relationship between different clients. Sattler et al.~\shortcite{sattler2020clustered} focused on a new setting where clients are adaptively partitioned into clusters, and a unique personalized model is trained for each cluster of clients.

Other personalized FL algorithms include carefully interpolating a model for each client. \ct{APFL}~\cite{deng2020adaptive} weights the global and the local model at each client. \ct{FedFomo}~\cite{zhang2020personalized} computes estimates of the optimal weights for each client's personalized model using a local validation set. \ct{FedAMP}~\cite{huang2021personalized} uses an attention-inducing function to compute an interpolated model as the prox-center for the personalized model.

\section{Adaptive Personalized Cross-Silo Federated Learning}
In this section, we look at personalized FL with more details, and present \ct{APPLE}, a framework for personalized cross-silo FL that adaptively learns to personalize the client models. Similar to most FL methods, in \ct{APPLE}, the training progresses in rounds. Each client iteratively downloads and uploads model parameters in each round. However, in \ct{APPLE}, each client uploads a constructing component of the \textbf{personalized model}, here called a \textbf{core model}. And the central server maintains the core models uploaded from the clients. Before we go into further details, we formulate the problem and define the notations which will be used throughout the paper.

\subsection{Problem Formulation}
In general, federated learning aims to improve model performance of individually trained models, by collaboratively training a model over a number of participating clients, without migrating the data due to privacy concerns. Specifically, the goal is to minimize:
\begin{equation}
    \min_{\vb*{w}} f_G(\vb*{w}) = \min_{\vb*{w}} \sum_{i=1}^N p_i F_i(\vb*{w}) ,
    \label{emr-G}
\end{equation}
where $f_G(\cdot)$ denotes the global objective. It is computed as the weighted sum of the $N$ local objectives, with $N$ being the number of clients and $p_i \geq 0$ being the weights. The local objective $F_i(\cdot)$ of client $i$ is often defined as the expected error over all data under local distribution $\mathcal{D}_i$, i.e. 
\begin{equation}
    F_i(\cdot)=\E_{\xi \sim \mathcal{D}_{i}}[\Ls(\cdot; \xi)] \approx \frac{1}{n_i}\sum_{\xi \in D_i^{tr}} \Ls(\cdot; \xi),
    \label{local-empirical}
\end{equation}
where $\xi$ represents the data under local distribution $\mathcal{D}_{i}$. As shown in Equation \ref{local-empirical}, $F_i(\cdot)$ is often approximated by the local empirical risk on client $i$ using its training set $D_i^{tr}$ ($n_i=|D_i^{tr}|$). The notable \ct{FedAvg} tries to solve the empirical risk minimization (EMR) by iteratively averaging the local updates of the copy of global model, i.e. $\vb*{w_{t+1}} = \sum_{i=1}^K p_i \vb*{w_{t+1}^i}$, where $K$ is the number of selected clients in each round, and the $p_i$'s are defined as the ratio between the number of data samples on client $i$, $n_i$, and the number of total data samples from all clients $n$, with constraints: $\forall i \in [N],\ p_i \geq 0$, and $\sum_i p_i = 1$.

In personalized FL, the global objective slightly changes to a more flexible form:

\begin{equation}
    \begin{split}
        \min_{\vb*{W}} f_P(\vb*{W}) & = \min_{\vb*{w_i}, i\in[N]} f_P(\vb*{w_1}, ...,\vb*{w_N})\\
        & = \min_{\vb*{w_i}, i\in[N]} \sum_{i=1}^N p_i F_i(\vb*{w_i}),
    \end{split}
    \label{emr-P}
\end{equation}
where $f_P(\cdot)$ is the global objective for the personalized algorithms, and $\vb*{W}=[\vb*{w_1},\vb*{w_2}, ...,\vb*{w_N}]$ is the matrix with all personalized models. In this work, we aim to obtain the optimal $\vb*{W^*} = \arg \min_{\vb*{W}} f_P(\vb*{W})$, which equivalently represents the optimal set of personalized models $\vb*{w_i^*}, i\in[N]$. In addition, we focus on the cross-silo setting of FL, which is differentiated with the cross-device setting by much smaller number of participating (stateful) clients, and no selection of clients are strictly needed at the beginning of each round.

\subsection{Adaptively Learning to Personalize}
As mentioned above, in \ct{APPLE}, each client uploads to the central server a \textbf{core model}, and downloads other clients' core models maintained on the server at the end and beginning of each round, respectively. In an ideal scenario where communication cost is not taken into account, each core model maintained on the server is downloaded by every client. In practice, communication is costly, and the limitations on the communication bandwidth always exist. We will discuss how \ct{APPLE} handles this in Section \ref{limited-comm}.

In \ct{APPLE}, after each client has downloaded the needed core models from the server, the personalized model for client $i$ is subsequently computed as

\begin{equation}
    \vb*{w_i^{(p)}} = \sum_{j=1}^N p_{i,j} \vb*{w_j^{(c)}},
    \label{apple-pers}
\end{equation}
where $\vb*{w_i^{(p)}}$ represents the personalized model of client $i$, and $\vb*{w_j^{(c)}}$ is the downloaded client $j$'s core model. Similar to some personalized FL algorithms that focus on interpolating a model for each client~\cite{acar2020federated,zhang2020personalized}, the personalized model here is also a convex combination of models. The difference is that in \ct{APPLE}, there is a unique set of learnable weights for each client. We use $p_{i,j}$ to denote the learnable weight on client $i$ for the downloaded core model $\vb*{w_j^{(c)}}$, and use $\vb*{p_{i}}=[p_{i,1},...,p_{i,N}]^T$ to denote the set of learnable weights on client $i$, calling it the \textbf{directed relationship (DR)} vector.

During local training on client $i$, after the personalized model $\vb*{w_i^{(p)}}$ is computed, we freeze the downloaded core models ($\vb*{w_j^{(c)}}, \ j \neq i$), and only update its local core model $\vb*{w_i^{(c)}}$ using a gradient-based method, such as local Stochastic Gradient Descent (SGD). Meanwhile, we adaptively update the DR vector, $\vb*{p_{i}}$, according to the local objective, i.e.
\begin{equation}
    \vb*{w_i^{(c)}} \leftarrow \vb*{w_i^{(c)}} - \eta_1 \frac{\partial}{\partial{\vb*{w_i^{(c)}}}} F_i(\vb*{w_i^{(p)}}),
    \label{apple-net-backprop}
\end{equation}

\begin{equation}
    \vb*{p_{i}} \leftarrow \vb*{p_{i}} - \eta_2 \frac{\partial}{\partial{\vb*{p_{i}}}} F_i(\vb*{w_i^{(p)}}).
    \label{apple-coef-backprop}
\end{equation}

Note that after a round of local training is finished, each client only uploads the local core model ($\vb*{w_i^{(c)}}$ for client $i$, $i\in [N]$) to the server. The DR vector $\vb*{p_i}$ is always maintained at client $i$ without any migration, which makes it impossible for others to infer the personalized model, and further protects the data privacy.

\subsection{Proximal Directed Relationships}
\label{proximal-dr}

In \ct{APPLE}, for each client, the learned global information is blended in the downloaded core models, whose contributions to the local personalized model are measured by the learnable weights in the DR vector. Ideally, the entry $p_{i,i}$, or ``self-relationship'', should be larger than the other entries in $\vb*{p_i}$, since the local core model $\vb*{w_i^{(c)}}$ is the only network trained with local distribution $\mathcal{D}_{i}$. On the other hand, for all $j \neq i$, $p_{i,j}$ should be somewhere in between $0$ and $p_{i,i}$, if the local personalized model $\vb*{w_i^{(p)}}$ can benefit more from $\vb*{w_j^{(c)}}$ (may happen if the distributions $\mathcal{D}_{i}$ and $\mathcal{D}_{j}$ are similar), while $p_{i,j}$ should be closer to $0$ or even negative, if $\vb*{w_j^{(c)}}$ results in potential negative transfer to $\vb*{w_i^{(p)}}$.

However, in a real-world situation, due to the data heterogeneity in FL, chances for similar distributions among clients are slim. Most off-diagonal entries in the DR matrix should be small. Without any constraint, this may result in a natural pitfall that the learned DR matrix is too quickly drawn to somewhere near the identity matrix (in terms of the Frobenius norm). This can lead the personalized models to hardly benefit from FL, and the training process undesirably resembles individual learning.

To address this issue and facilitate collaboration between clients, we penalize the directed relationship by adding a proximal term~\cite{rockafellar1976monotone} to the local empirical risk, which is summarized in \ct{APPLE} in Equation \ref{apple-local-empirical}.

\begin{equation}
    F_i(\vb*{w_i^{(p)}})= \frac{1}{n_i}\sum_{\xi \in D_i^{tr}} \Ls(\vb*{w_i^{(p)}}; \xi)  + \lambda(r) \frac{\mu}{2} || \vb*{p_i}-\vb*{p_0}||_2^2 .
    \label{apple-local-empirical}
\end{equation}
In Equation \ref{apple-local-empirical}, $\lambda$ is a dynamic function ranging from $0$ and $1$, with respect to the round number, $r$, and $\mu$ is a scalar coefficient for the proximal term. The prox-center is at $\vb*{p_0}=[n_1/n, ..., n_N/n]$. It is obvious that \ct{FedAvg} is a special case of \ct{APPLE} by setting $\mu$ to $\infty$, which infers that a larger coefficient of the proximal term can push the personalized model to a global model, facilitating collaboration between clients. While this may benefit the personalized model to learn high-level features, it is not always desired throughout the training. Ultimately, with the learned high-level features, the personalized models should still focus on how to be personalized.

\begin{algorithm}[tb]
    \caption{\ct{APPLE}}
    \label{apple-alg}
    \textbf{Input}:  $N$ clients, learning rates $\eta_1, \eta_2$, number of total rounds $R$, proximal term coefficients $\lambda(r), \mu$, prox-center $\vb*{p_0}$\\
    \textbf{Output}: Personalized models $\vb*{w_1^{(p)}},\vb*{w_2^{(p)}},...,\vb*{w_N^{(p)}}$ on site of corresponding client.
    \begin{algorithmic}[1] 
        \STATE $\forall i \in [N]$, randomly initialize core model $\vb*{w_{i}^{(c)}}$ on server
        \STATE $\forall i \in [N]$ \textbf{in parallel}, initialize local DR vector $\vb*{p_i}$
        \FOR{$r \leftarrow 1,2,...,R$}
            \FOR{$i \leftarrow 1,2,...,N$ \textbf{in parallel}}
                \STATE Download core models from server as needed.
                \STATE Iteratively optimize the local core model $\vb*{w_i^{(c)}}$ and local DR vector $\vb*{p_i}$ by the following:
                    \STATE $\text{    } \ \ \ \ $Compute personalized model $\vb*{w_i^{(p)}}$ by Eq. (\ref{apple-pers})
                    \STATE $\text{    } \ \ \ \ $Compute empirical risk $F_i(\vb*{w_i^{(p)}})$ by Eq. (\ref{apple-local-empirical})
                    \STATE $\text{    } \ \ \ \ $Update $\vb*{w_i^{(c)}}$ and DR vector $\vb*{p_i}$ by Eqs. (\ref{apple-net-backprop}, \ref{apple-coef-backprop})
                \STATE When optimization is finished, upload local core model $\vb*{w_{i}^{(c)}}$ to the server
            \ENDFOR
        \ENDFOR
        \STATE \textbf{return} $\vb*{w_1^{(p)}},\vb*{w_2^{(p)}},...,\vb*{w_N^{(p)}}$
    \end{algorithmic}
\end{algorithm}

To this end, we design the function $\lambda(\cdot)$ with a certain type of decay in terms of the round number, $r$, inspired by Wang et al.~\shortcite{wang2019dynamic}, and call such $\lambda(r)$ a \textit{loss scheduler}. More details regarding the loss scheduler are presented in Appendix \ref{apdx-loss-scheduler}. We summarize the steps of \ct{APPLE} in Algorithm \ref{apple-alg}.

\subsection{\ct{APPLE} under Limited Communication Budget\label{limited-comm}}
The collaboration between clients in \ct{APPLE} relies on the downloaded core models at the beginning of each round. Without considering communication limitations, each client can download all other clients' core models, which maximizes the potential pair-wise collaboration between clients.

However, although the number of clients in cross-silo settings will not be as large compared to the cross-device settings, the communication cost of downloading all core models to each client is still considerable. While this issue can be mitigated by techniques including quantization~\cite{xu2018deep} and knowledge distillation~\cite{hinton2015distilling}, in the worst-case scenario, the communication per round still cost $N$ times more overhead than algorithms that only download one model for each client per round (e.g. \ct{FedAvg}).

\label{download-which}To address this issue, we restrict the number of models a client can download per round, denoted by $M$. Under limited communication budget ($1 \leq M \leq N-1$), briefly, \ct{APPLE} decides which $M$ core models to download to each client by the following rules: on client $i$, the core model of client $j$ will be downloaded if it has never been downloaded on client $i$ (breaks tie randomly); if all other clients' core models have all been downloaded at least once, with high probability, priority goes to client $j$'s core model who has a large $p_{i,j}$. We elaborate this process in Appendix \ref{apdx-core-model-selection}.

\section{Experiments}
\label{experiments}
In this section, we demonstrate the effectiveness of \ct{APPLE} with experiments under two different non-IID federated settings. We show the empirical convergence behaviors and the generalization performances with respect to each client on different image datasets. In addition, we study the transition of the pairwise directed relationships between different clients throughout the training process. Last but not least, we also investigate the performance of \ct{APPLE} under different levels of limited communication budget.

To evaluate the convergence behavior, we plot the training losses and test accuracies against the number of trained rounds. For the personalized methods, the training loss and test accuracy are computed in a way such that if a data sample resides on client $i$, then we use the personalized model of client $i$ to conduct inference with it. In addition, we quantify the performance of the methods by computing the test accuracies with respect to each client, and the \label{BMCTA}\textit{best mean client test accuracy (BMCTA)} (best over all rounds, mean over all clients), a metric also used by Huang et al.~\shortcite{huang2021personalized}.

\subsection{Experimental Setup}

\begin{figure*}[t]
    \centering
    \includegraphics[width=\textwidth]{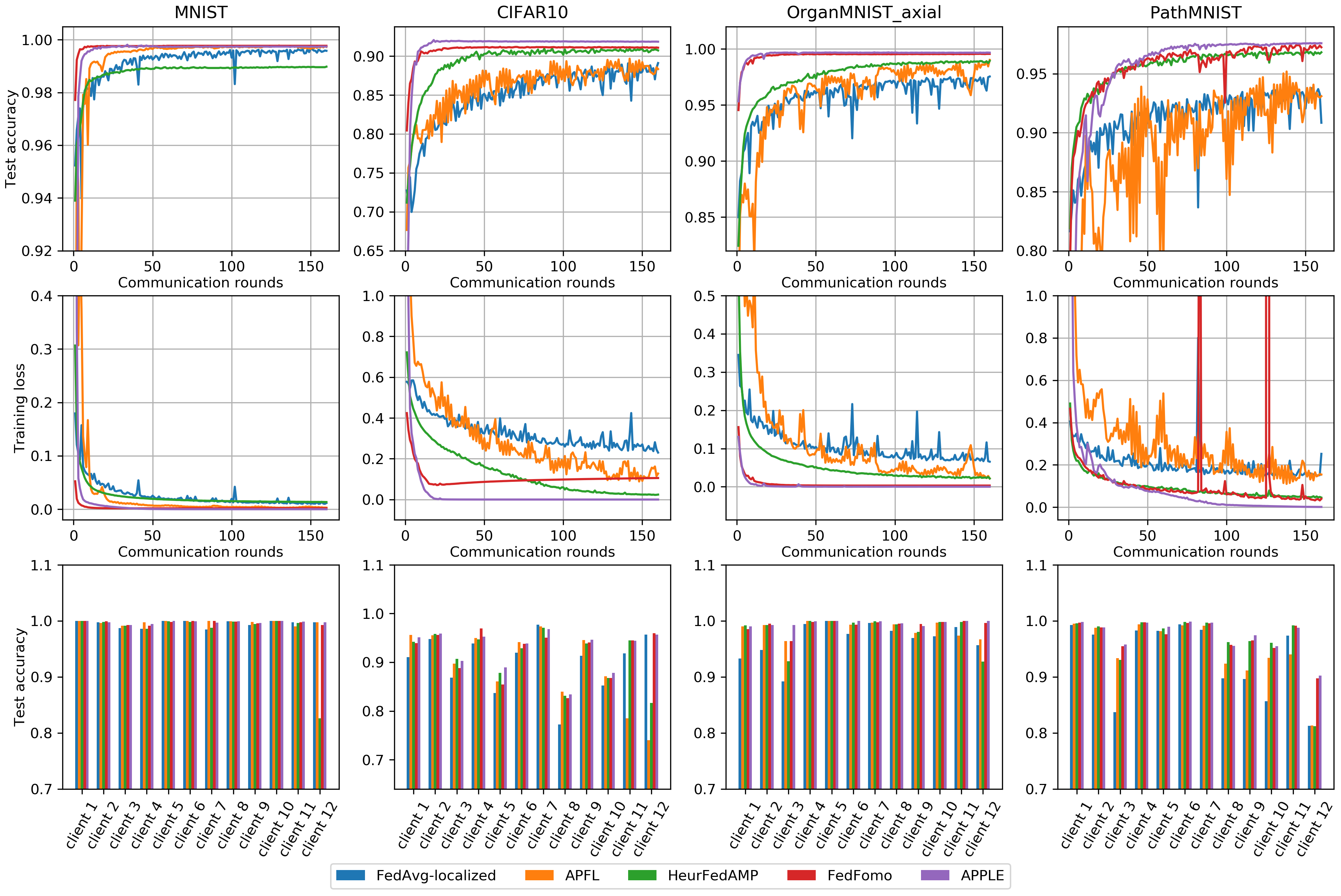}
    \caption{The training loss, test accuracy and client test accuracies of personalized methods under the pathological non-IID setting.} \label{res:pic-path}
\end{figure*}

\paragraph{Datasets.}
We use four public datasets including two benchmark datasets: MNIST and CIFAR10, and two medical imaging datasets from the MedMNIST datasets collection~\cite{yang2021medmnist}, namely the OrganMNIST(axial) dataset: an 11-class of liver tumor image dataset, and the PathMNIST dataset: a 9-class colorectal cancer image dataset. We partition each dataset into a training set and a test set with the same distributions (if such split does not pre-exist). Then we transform the datasets according to a non-IID distribution, and ensure the same distributions of local training and local test set.


\paragraph{Pathological and Practical Non-IID Settings.}\label{non-iid-description} We design two non-IID distributions for empirical evaluation, namely the \textit{pathological non-IID} and the \textit{practical non-IID}. For the pathological non-IID, we follow precedent work and select two random classes for each client. A random percentage of images from each of the two selected classes is assigned to the client. To simulate a cross-silo setting, the number of clients is set to be 12. For the practical non-IID, our endeavor aims to simulate a more realistic cross-silo federation of medical institutes. To this end, we partition each class of the dataset into 12 shards (corresponding to the 12 clients): 10 shards of 1\%, 1 shard of 10\% and 1 shard of 80\% images within this class. A randomly picked shard from each class is assigned to each client, so that every client will possess data from every class. The practical non-IID setting is more similar to the real-world FL in medical applications. This is because the datasets at medical institutes most likely contain a variety of categories of data. And due to different demographic distributions of patients, medical institutes located in different regions may have more frequent occurrences of different categories of data. As a result, these datasets are often imbalanced with different majority classes, and vary largely in size. Appendix \ref{apdx-datasets} shows the data distributions in further details.

\begin{table}[t]
    \centering
    \begin{tabular}{wl{2.3cm}M{1cm}M{1cm}M{1cm}M{1cm}}
    \toprule
    \makecell{\textbf{\small{Pathological}}\\\textbf{\small{non-IID}}}&\small{MNIST}&\small{CIFAR10}&\small{Organ-\newline{MNIST}\newline{(axial)}} &\small{Path-\newline{MNIST}}\\ \midrule
    \ct{Separate} & 97.34 & 74.96 & 93.14 & 87.09\\
    \ct{FedAvg} & 95.71 & 51.44 & 59.43 & 56.61\\ 
    \ct{FedAvg-local} & 99.52 & 90.10 & 96.76 & 93.21\\
    \ct{FedAvg-FT} & 99.43 & 90.49 & 97.03 & 92.31\\
    \ct{FedProx-FT} & 99.43 & 90.49 & 97.03 & 92.38\\
    \ct{APFL} & 99.75 & 89.30 & 98.72 & 94.98\\
    \ct{HeurFedAMP} & 98.13 & 91.10 & 98.39 & 96.55\\
    \ct{FedFomo}  & 99.71 & 91.96 & 99.31 & 97.24\\ \midrule
    \ct{APPLE}, $\mu=0$ & 99.73 & 92.22 & \textbf{99.66} & 96.78\\
    \ct{APPLE}, $\mu \neq 0$ & \textbf{99.77} & \textbf{92.68} & 99.61 & \textbf{97.51}\\
    \midrule \midrule
    \makecell{\textbf{\small{Practical}}\\\textbf{\small{non-IID}}}&\small{MNIST}&\small{CIFAR10}&\small{Organ-\newline{MNIST}\newline{(axial)}} &\small{Path-\newline{MNIST}}\\ \midrule
    \ct{Separate} & 78.20 & 63.06 & 65.21 & 61.36 \\
    \ct{FedAvg} & 94.00 & 34.32 & 86.56 & 53.83 \\ 
    \ct{FedAvg-local} & 97.47 & 71.99 & 93.75 & 78.70\\
    \ct{FedAvg-FT} & 97.66 & 72.08 & 94.13 & 78.69 \\
    \ct{FedProx-FT} & 97.66 & 72.08 & 94.13 & 78.69\\
    \ct{APFL} & 98.80 & 71.19 & 95.53 & 86.35 \\
    \ct{HeurFedAMP} & 97.45 & 69.54 & 86.82 & 79.33\\
    \ct{FedFomo} & 98.05 & 70.15 & 82.86 & 79.39\\ \midrule
    \ct{APPLE}, $\mu=0$ & \textbf{99.00} & 75.62 & \textbf{95.70} & 84.22\\
    \ct{APPLE}, $\mu \neq 0$ & 98.97 & \textbf{77.41} & 95.62 & \textbf{86.39}\\
    \bottomrule
    \end{tabular}
    \caption{\hyperref[BMCTA]{Best mean client test accuracy (BMCTA)} of the four datasets under the pathological and practical non-IID settings. Highest performance is represented in bold.}\label{res:tab-perf}
\end{table}

\paragraph{Compared Baselines.} We compare \ct{APPLE}, with and without proximal DR penalty, against the following approaches: (1) \ct{Separate} training, meaning the clients' models are trained purely locally without FL; (2) the \ct{FedAvg}~\cite{mcmahan2017communication} which takes the average of the locally trained copies of the global model; (3) \ct{FedAvg-local}, a na\"ive personalized approach of using the locally trained copy of \ct{FedAvg}'s global model; (4) a fine-tuning approach~\cite{wang2019federated} on the \ct{FedAvg} and on the \ct{FedProx}~\cite{li2020federated}, here denoted as \ct{FedAvg-FT} and \ct{FedProx-FT}, respectively; (5) \ct{APFL}~\cite{deng2020adaptive}, a personalized method using a mixture of the global and the local model; (6) \ct{HeurFedAMP}~\cite{huang2021personalized}, a personalized method on the cross-silo setting with federated attentive message passing; and (7) \ct{FedFomo}~\cite{zhang2020personalized}, a personalized method that computes first-order approximations for the personalized models. We train each method 160 rounds with 5 local epochs and summarize the results as follows.

\subsection{Experimental Results}
\label{exp:res}
We summarize the empirical convergence behavior and performance under the pathological and the practical non-IID settings in Figure \ref{res:pic-path}, Table \ref{res:tab-perf} and Figure \ref{res:pic-prac} (in Appendix \ref{apdx-prac-pic}). Across all datasets and non-IID settings, our proposed method has fast convergence behaviors, and achieves highest BMCTAs. Specifically, under the pathological non-IID setting, the separate training reaches comparable performance with other methods, due to little similarity in data distribution shared by different clients, and the small number of classes in each client. With a direct averaging of the local updates as in \ct{FedAvg}, and no fine-tuning as in \ct{FedAvg-FT} and \ct{FedProx-FT}, the global model is hardly able to boost the performance of separate training. The personalized FL methods bring further improvement to the na\"ive personalization, and \ct{APPLE} outperforms the other compared personalized FL methods. For the practical non-IID setting, since the local data distributions across all clients are more complicated in terms of the number of classes, and local majority classes, a high performance requires more careful integration of the global information. As a result, the fine-tuning methods outperform some personalized methods, and \ct{APPLE} also reaches state-of-the-art performance in all settings.

\begin{table}[t]
    \centering
    \begin{tabular}{wl{0.8cm}M{1.6cm}M{0.91cm}M{0.91cm}M{0.91cm}M{0.91cm}}
    \toprule
    \multicolumn{2}{c}{\makecell{\textbf{\small{Pathological}}\\\textbf{\small{non-IID}}}}&\small{MNIST}&\small{CIFAR10}&\small{Organ-\newline{MNIST}\newline{(axial)}} &\small{Path-\newline{MNIST}}\\ \midrule
    \multirow{2}{*}{$M=11$} & \ct{FedFomo} & 99.71 & 91.96 & 99.31 & 97.24 \\
    & \ct{APPLE} & 99.73 & 92.22 & 99.66 & 96.78\\ \midrule
    \multirow{2}{*}{$M=7$} & \ct{FedFomo} & 99.71 & 91.95 & 99.31 & 97.33\\
    & \ct{APPLE} & 99.73 & 92.17 & 99.53 & 97.15\\ \midrule
    \multirow{2}{*}{$M=5$} & \ct{FedFomo} & 99.71 & 91.94 & 99.31 & 97.40\\
    & \ct{APPLE} & 99.72 & 92.28 & 99.48 & 97.17 \\ \midrule
    \multirow{2}{*}{$M=2$} & \ct{FedFomo} & 99.71 & 91.98 & 99.31 & 97.25\\
    & \ct{APPLE} & 99.70 & 92.41 & 99.47 & 97.11\\ \midrule
    \multirow{2}{*}{$M=1$} & \ct{FedFomo} & 99.71 & 91.95 & 99.31 & 97.15\\
    & \ct{APPLE} & 99.66 & 92.31 & 99.59 & 96.29\\
    \midrule \midrule
    \multicolumn{2}{c}{\makecell{\textbf{\small{Practical}}\\\textbf{\small{non-IID}}}}&\small{MNIST}&\small{CIFAR10}&\small{Organ-\newline{MNIST}\newline{(axial)}} &\small{Path-\newline{MNIST}}\\ \midrule
    \multirow{2}{*}{$M=11$} & \ct{FedFomo} & 98.05 & 70.15 & 82.86 & 79.39 \\
    & \ct{APPLE} & 99.00 & 75.62 & 95.70 & 84.22 \\ \midrule
    \multirow{2}{*}{$M=7$} & \ct{FedFomo} & 97.65 & 70.24 & 80.88 & 80.19 \\
    & \ct{APPLE} & 98.70 & 76.14 & 94.21 & 84.07 \\ \midrule
    \multirow{2}{*}{$M=5$} & \ct{FedFomo} & 97.47 & 70.44 & 82.83 & 79.62\\
    & \ct{APPLE} & 98.45 & 75.63 & 94.49 & 85.46 \\ \midrule
    \multirow{2}{*}{$M=2$} & \ct{FedFomo} & 96.51 & 69.87 & 79.53 & 79.26 \\
    & \ct{APPLE} & 98.29 & 74.84 & 92.29 & 84.64 \\ \midrule
    \multirow{2}{*}{$M=1$} & \ct{FedFomo} & 91.54 & 69.93 & 78.37 & 75.17 \\
    & \ct{APPLE} & 98.52 & 73.03 & 93.55 & 83.35 \\ 
    \bottomrule
    \end{tabular}
    \caption{BMCTA of \ct{APPLE} ($\mu=0$) and \ct{FedFomo} with a maximum of $M$ models ($N=12$) to download for each client per round under the pathological non-IID setting.}\label{res:limit-comm}
\end{table}

Next, we report the performance of \ct{APPLE} under limited communication budget. With the same levels of communication restriction, we compare \ct{APPLE} against \ct{FedFomo} since \ct{FedFomo} also needs to download $N-1$ models to each client by default. We restrict the maximum number of downloaded models for each client per round (denoted by $M$) to different levels ($M=11, 7, 5, 2, 1$). Table \ref{res:limit-comm} shows the results under these settings. For the pathological non-IID setting, results are mixed across different datasets. \ct{APPLE} outperforms \ct{FedFomo} on the CIFAR10 and OrganMNIST (axial) dataset, while \ct{FedFomo} reaches higher performance on the PathMNIST dataset. For the practical non-IID setting, \ct{APPLE} outperforms \ct{FedFomo} across all datasets and different levels of limitations. Note that in \ct{APPLE}, less downloaded models (smaller $M$) do not necessarily lead to an inferior performance. This is because \hyperref[download-which]{the proposed rule of picking which models to download} tends to download to a client the top $M$ core models that have the highest chance to benefit the client.


\begin{figure*}[t]
    \centering
    \includegraphics[width=\textwidth]{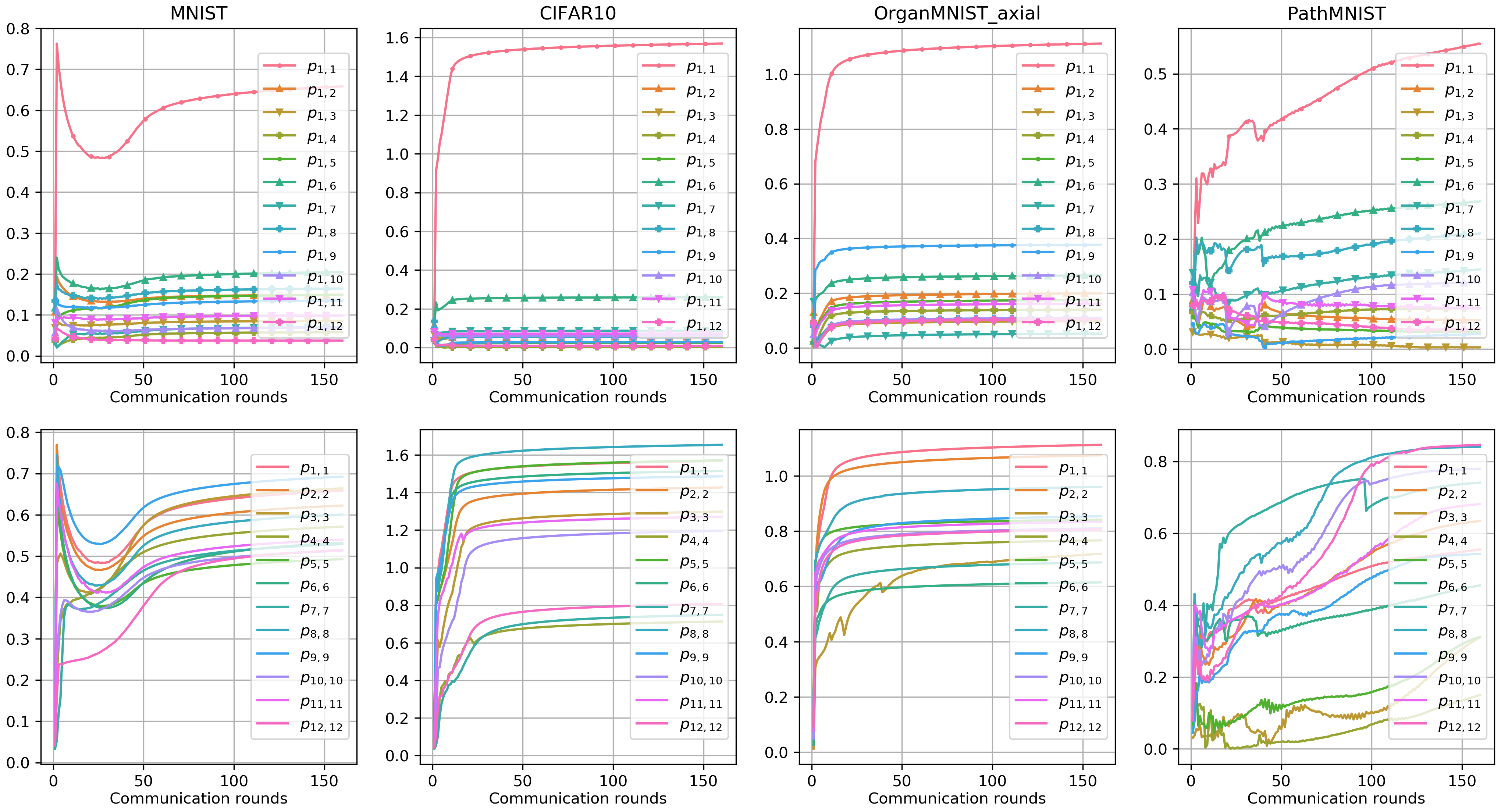}
    \caption{Directed Relationships of different datasets under the pathological non-IID setting. The first row shows the DRs on client $1$. The second row shows the ``self-relationships'', $p_{i,i}$, for each client.} \label{dr:path}
\end{figure*}

Furthermore, we visualize the change of the directed relationships throughout the training process. Specifically, we study the local DRs on client $1$, i.e. $p_{1,i}$'s, and the $p_{i,i}$'s for all clients. The visualizations are shown in Figure \ref{dr:path} and Figure \ref{dr:prac} (in Appendix \ref{apdx-vis-dr}) for the trajectories of DRs under the pathological and the practical non-IID settings, respectively. As mentioned in Section \ref{proximal-dr}, the self-relationship, $p_{i,i}$, should be larger than the other local DRs, since on client $i$, the only updated core model is $\vb*{w_{i}^{(c)}}$. And depending on the similarity between the data distributions on client $i$ and on client $j$, larger similarity will push $p_{i,j}$ towards $|p_{i,i}|$ and lower similarity will push $|p_{i,j}|$ towards $0$. Figure \ref{dr:path} empirically shows these properties. For instance, in CIFAR10, even in the same class, images can have high variance, different client shares little similarity in distribution, so the $p_{i, i}$'s are large and $p_{i, j}$'s are small. In PathMNIST, $p_{1,6}$ and $p_{1,8}$ are closer to $p_{1,1}$, which makes sense since client $1$ and $6$ both have a large portion of images from class $1$, and client $1$ and $8$ both have a large portion of images from class $7$ (refer to Figure \ref{data:path} in Appendix \ref{apdx-datasets} for the data distribution).

\section{Conclusions}
In this work, we proposed \ct{APPLE}, a novel personalized cross-silo federated learning framework for non-IID data, that adaptively learns the quantification of how much each client can benefit from other clients’ models, i.e. the directed relationships (DRs). We introduced a proximal DR penalty to control the training between global and local objectives. In addition, we evaluated our method's convergence and generalization behaviors on four image datasets under two non-IID settings. The results showed the overall superior effects of \ct{APPLE} over several related personalized FL alternatives. Through the visualization of the DRs, our study empirically shows that \ct{APPLE} enables clients to adaptively take more advantage from other clients with similar distributions, while mitigating the potential non-beneficial influences or negative transfers from clients with drastically different data distributions. We also demonstrated that under limited communication budget, \ct{APPLE} can still reach state-of-the-art performance. As future work, we plan to further evaluate our algorithm's robustness with real-world medical datasets.

\section*{Acknowledgements}
This work was supported in part by a National Institutes of Health (NIH) / National Cancer Institute (NCI) grant (1R01CA218405), a National Science Foundation (NSF) grant (CICI: SIVD: 2115082), the grant 1R01EB032896 as part of the NSF/NIH Smart Health and Biomedical Research in the Era of Artificial Intelligence and Advanced Data Science Program, a Pitt Momentum Funds scaling award (Pittsburgh Center for AI Innovation in Medical Imaging), and an Amazon AWS Machine Learning Research Award. This work used the Extreme Science and Engineering Discovery Environment (XSEDE), which is supported by NSF grant number ACI-1548562. Specifically, it used the Bridges-2 system, which is supported by NSF award number ACI-1928147, at the Pittsburgh Supercomputing Center.

\bibliographystyle{named}
\bibliography{ijcai22}


\begin{figure*}[ht!]
    \centering
    \includegraphics[width=0.9\textwidth]{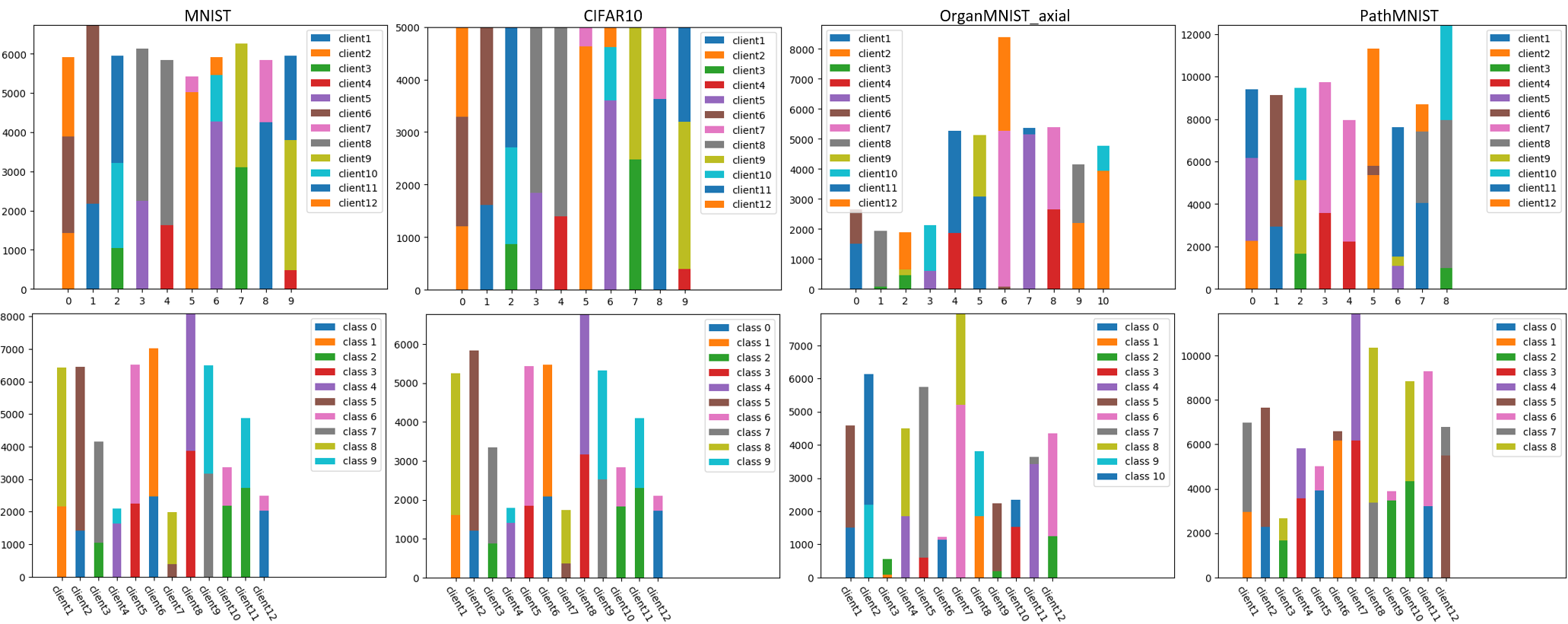}
    \caption{Data distribution of the pathological non-IID setting. The first row represents which clients the data is assigned to. The second row represents the local label distribution of each client.} \label{data:path}
\end{figure*}

\begin{figure*}[h!]
    \centering
    \includegraphics[width=0.9\textwidth]{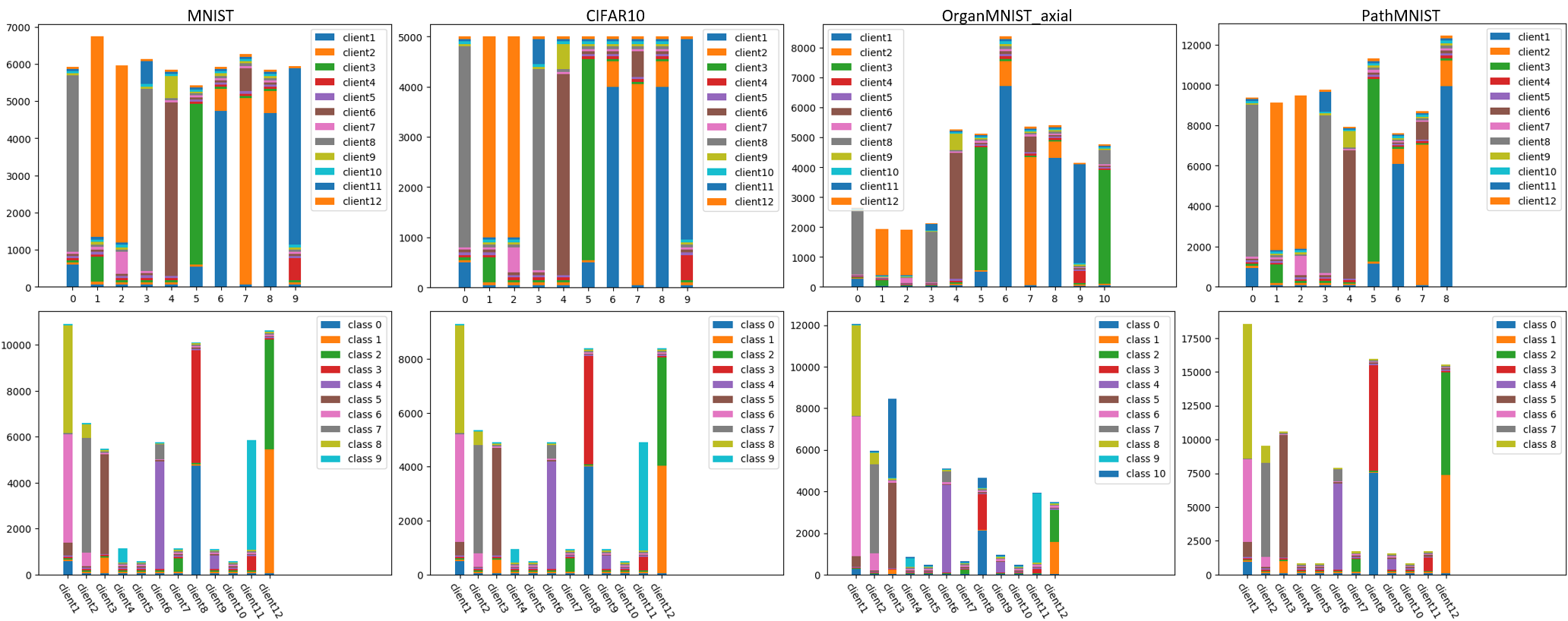}
    \caption{Data distribution of the practical non-IID setting. The first row represents which clients the data is assigned to. The second row represents the local label distribution of each client.} \label{data:prac}
\end{figure*}


\appendix


\section{Details of the Algorithm Design}
\label{apdx-alg-design}

\begin{figure*}[ht!]
    \centering
    \includegraphics[width=\textwidth]{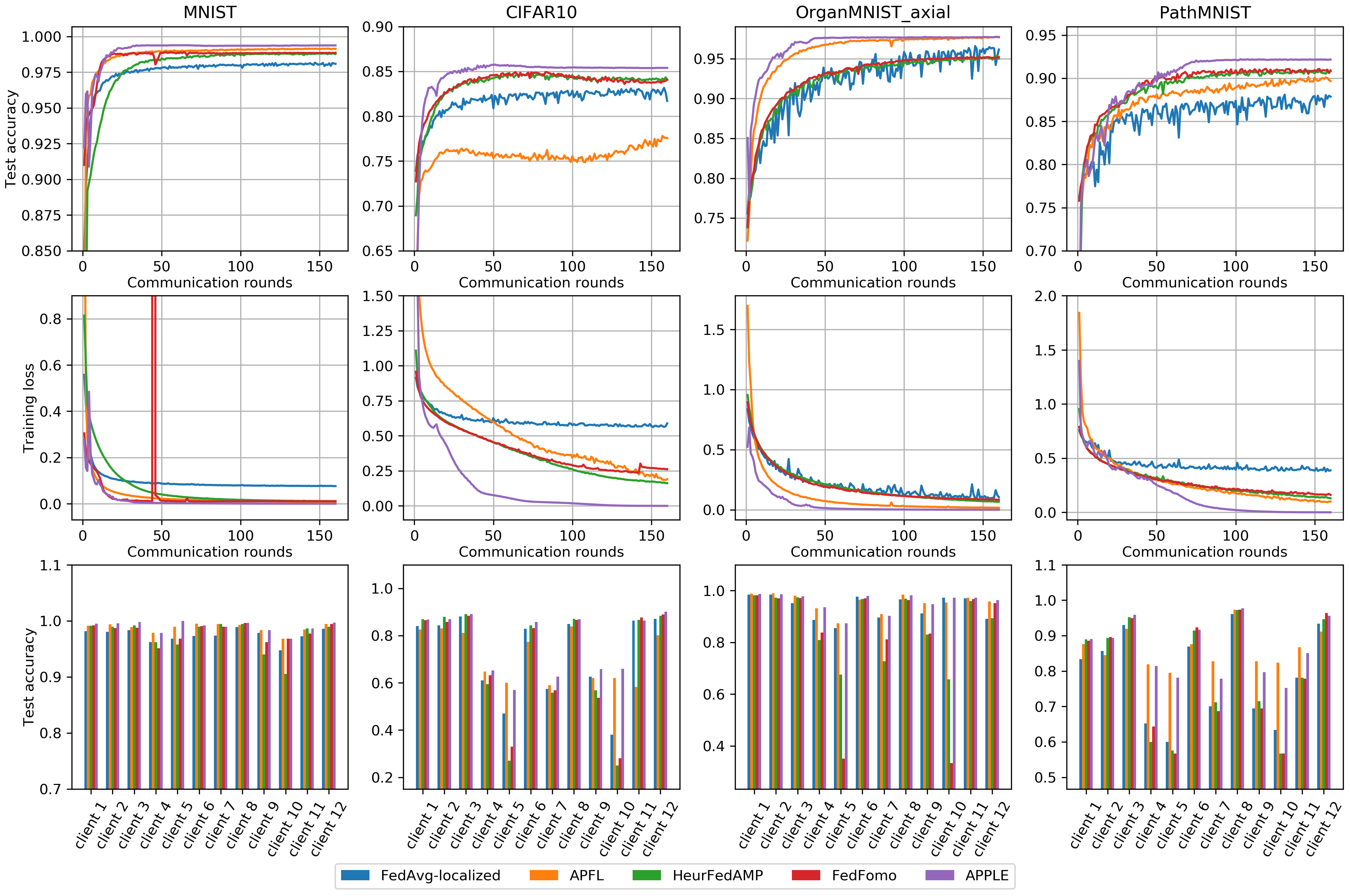}
    \caption{The training loss, test accuracy and client test accuracies of personalized methods under the practical non-IID setting.} \label{res:pic-prac}
\end{figure*}

\subsection{Loss Scheduler for Proximal Directed Relationships}
\label{apdx-loss-scheduler}
To push the model to learn more global information at the beginning of the training while gradually transitioning to focusing more on local training, we design the loss scheduler, $\lambda(r)$, to be a monotonically decreasing function between $1$ and $0$ in terms of the current training round $r$. Theoretically, $\lambda(r)$ can be designed in different forms, as long as it has the above property. Here, we explore the following two types of loss scheduler (shown in Figure \ref{loss-scheduler-pic}) and treat the choice of loss scheduler type as an additional hyperparameter. In both of the following two loss scheduler expressions, $L$ is the round number after which the loss scheduler's value is always $0$:
\begin{figure}[H]
    \centering
    \includegraphics[width=0.5\textwidth]{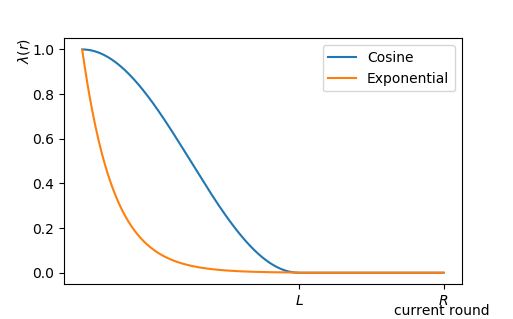}
    \caption{Loss scheduler} \label{loss-scheduler-pic}
\end{figure}

\begin{itemize}
    \item a cosine-shaped scheduler: $\lambda(r)=\left(\cos(r\pi / L)+1 \right) / 2$ indicating the learning focus transitions gradually from global to local.
    \item an exponentially decreasing scheduler: $\lambda(r)=\epsilon^{r/L}, \ \epsilon=10^{-3}$, indicating a rapid transition from global to local.
\end{itemize}

\subsection{Core Model Selection under Limited Communication Budget}
\label{apdx-core-model-selection}
Under the constraint that a maximum of $M$ core models can be downloaded for each of the $N$ clients per round, we first compute a normalized set of powers as the probabilities for the core models to be selected. The base of the powers is shared among all clients, and is computed by
\begin{equation}
    b(r) = \max(1.5, rM/N),
\end{equation} 
where $b(r)$ is the base with respect to the current training round $r$, and $rM/N$ computes the mean downloaded times per core model for the first $r$ rounds. For client $i$, the exponent of the powers are $|p_{i,j}|$'s. In other words, the core model $\vb*{w_{j}^{(c)}}$ will be downloaded to client $i$ with probability:

\begin{equation}
    P(\vb*{w_{j}^{(c)}} \text{downloaded to client }i)=\frac{b(r)^{|p_{i, j}|}}{\sum_{j=1,j\neq i}^N b(r)^{|p_{i, j}|}}
\end{equation}

The reasoning behind this exponential design is that, as training progresses and $r$ increases, $|p_{i,j}|$ gradually represents the contribution of core model $\vb*{w_j^{(c)}}$ on client $i$ with more \textit{confidence}. However, for the first several rounds (with small $r$) where the core model $\vb*{w_j^{(c)}}$ still has a large potential to update, the confidence of $|p_{i,j}|$ to represent the contribution is still small. With an exponential design, where the base is correlated to the mean downloaded times per core model, this growth in confidence can be better represented.

\section{Experiments}
\subsection{Datasets}
\label{apdx-datasets}

We partition each dataset into pathological and practical non-IID distributions. Figure \ref{data:path} and Figure \ref{data:prac} shows the partition of the training set with respect to ``where do the images of each class go'' and ``what is the label distribution on each client''. For example Figure \ref{data:path} bottom right plot (PathMNIST dataset) shows that for the PathMNIST dataset under the pathological non-IID setting, client $1$ and client $6$ both contain a large portion of data from class $1$, which explains why the visualization in Figure \ref{dr:path} for the PathMNIST dataset demonstrates that $p_{1,6}$ and $p_{1,8}$ are closer to $p_{1,1}$ than other DRs.

\begin{figure*}[ht]
    \centering
    \includegraphics[width=\textwidth]{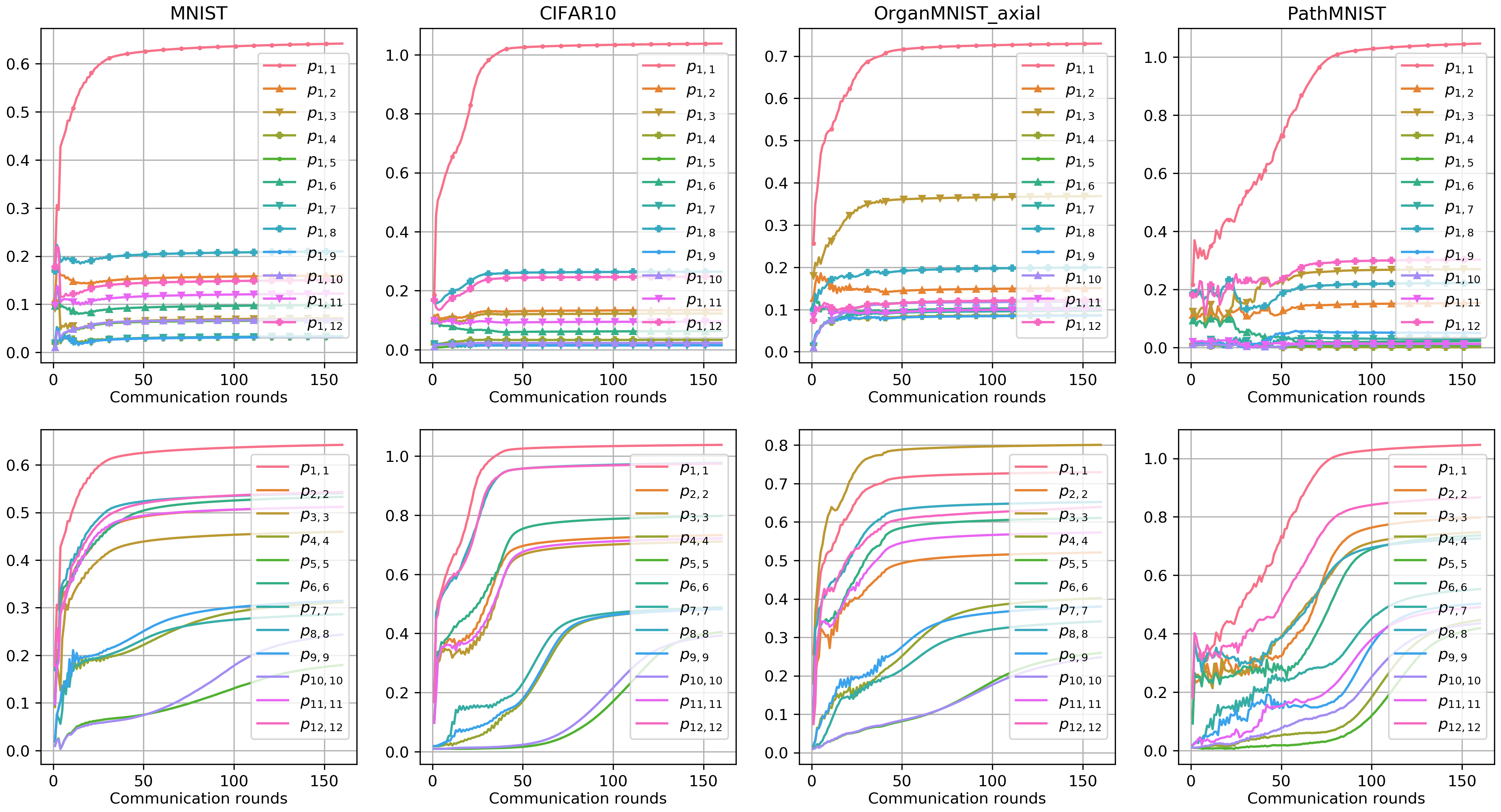}
    \caption{Directed Relationships of different datasets under the practical non-IID setting. The first row shows the DRs on client $1$. The second row shows the ``self-relationships'', $p_{i,i}$, for each client.} \label{dr:prac}
\end{figure*}

\subsection{Empirical Convergence Behavior under the Practical Non-IID Setting}
\label{apdx-prac-pic}

We show the experimental results of the practical non-IID setting in Figure \ref{res:pic-prac}. Under this setting, the importance of learning global information will be increased due to the following major aspects of the data: the datasets on the clients (1) contain a variety of categories, increasing the variance of the data; (2) are likely to be imbalanced with a different majority class on different clients; (3) are largely different in size, and local training on a small dataset might quickly overfit. Consequently, personalized methods that address a careful global information update of the model, such as \ct{APPLE} and \ct{FedFomo}, have natural advantages in this regard. Experimental results in Figure \ref{res:pic-prac} show the advantages of \ct{APPLE} and \ct{FedFomo} over other compared personalized methods, and \ct{APPLE} and \ct{FedFomo} achieve similarly fast convergence.

\subsection{Visualization of Directed Relationships Throughout Training}
\label{apdx-vis-dr}

We visualize the trajectories of the directed relationship vectors under the practical non-IID setting in Figure \ref{dr:prac}. Under the practical non-IID setting, since a large portion of samples in each class are assigned to only one client (recall the $80\% \times 1$, $10\% \times 1$, $1\% \times 10$ split described in Section \ref{non-iid-description}), less can be inferred about the DRs given each client's data distribution. We elaborate this through an example of \ct{FedFomo}. \ct{FedFomo} is a personalized FL method that focuses on weighting the personalized models with local validation sets. As the majority class on a different client is different, the personalized model on client $j$ can hardly perform well on client $i$. This results in less weight for client $j$'s personalized model on client $i$, and it fails to maximize the global information that can be learned. Our proposed method, \ct{APPLE}, takes a different scheme from \ct{FedFomo}. Rather than deciding the weights of other clients' personalized models purely based on the validation performance, \ct{APPLE}'s learnable DRs prevent the waste of other clients' core models. The learnable DRs enable to adaptively optimize the joint contribution from each downloaded core model. This is empirically demonstrated in Figure \ref{dr:prac} (in the OrganMNIST (axial) dataset). Although client $1$ and client $8$ do not share any majority class (refer to Figure \ref{data:prac}), $p_{1,8}$ can still be large, as long as the personalized model learns a beneficial assignment of each downloaded core model's contribution.

\subsection{Additional Implementation Details}
We used the pre-existing training set and test set of MNIST  and CIFAR10. For the two datasets from MedMNIST, since the training and test datasets are different in terms of the distribution, we combined them and split it into a new training set of $80\% $ of the entire dataset, and a new test set of the remaining $20\% $.

We adopted the classic four-layer CNN model. The model has two $5 \times 5$ convolutional layers followed by a fully connected layer with 500 units and another fully connected layer with the number of units equals to the number of classes. 

For each method, we trained the model for 160 rounds of 5 local epochs using a batch size of 256. We used SGD as the optimizer with 0.9 momentum, and chose the best performing learning rate in $\{10^{-2}, 10^{-3}, 10^{-4}\}$, and learning rate decay in $\{1.0, 0.9964 \ (=\sqrt[100]{0.7}), 0.9\}$. For \ct{APPLE}, we selected the loss scheduler type from cosine and exponential, $\mu$ from $\{ 1.0, \ 0.1,\ 0.01 \}$, and $L$ from $\{10\%, 20\%, 30\% \}$ of the number of total training rounds. The detailed hyperparameter values are summarized in Table \ref{apple-hyperparameters}

\begin{table}[htp!]
    \centering
    \begin{tabular}{wl{2.45cm}M{0.85cm}M{0.95cm}M{0.95cm}M{0.95cm}}
    \toprule
    \small{Pathological non-IID}&\small{MNIST}&\small{CIFAR10}&\small{Organ-\newline{MNIST}\newline{(axial)}} &\small{Path-\newline{MNIST}}\\ \midrule
    Net's learning rate & $10^{-2}$ & $10^{-2}$ & $10^{-2}$ & $10^{-2}$ \\
    DRs' learning rate & $10^{-3}$ & $10^{-3}$ & $10^{-3}$ & $10^{-4}$\\
    Loss scheduler type & cos & exp. & exp. & cos\\
    $\mu$ & 0.1 & 0.001 & 0.001 & 1.0 \\
    $L$ & 30\% & 20\% & 20\% & 10\% \\
    \midrule \midrule
    \small{Practical non-IID}&\small{MNIST}&\small{CIFAR10}&\small{Organ-\newline{MNIST}\newline{(axial)}} &\small{Path-\newline{MNIST}}\\ \midrule
    Net's learning rate & $10^{-2}$ & $10^{-2}$ & $10^{-2}$ & $10^{-2}$\\
    DRs' learning rate & $10^{-3}$ & $10^{-4}$ & $10^{-4}$ & $10^{-4}$\\
    Loss scheduler type & cos & cos & cos & cos\\
    $\mu$ & 0.01 & 0.001 & 0.1 & 1.0 \\
    $L$ & 30\% & 20\% & 20\% & 10\% \\
    \bottomrule
    \end{tabular}
    \caption{The hyperparameter values used in \ct{APPLE}} \label{apple-hyperparameters}
\end{table}

\end{document}

%% file: math_commands.tex
\usepackage{amsmath,amsfonts,bm}

\def\eqref#1{equation~\ref{#1}}

\def\1{\bm{1}}

\def\vb{{\bm{b}}}

\DeclareMathAlphabet{\mathsfit}{\encodingdefault}{\sfdefault}{m}{sl}
\SetMathAlphabet{\mathsfit}{bold}{\encodingdefault}{\sfdefault}{bx}{n}

\newcommand{\E}{\mathbb{E}}
\newcommand{\Ls}{\mathcal{L}}

